\documentclass{article}
\usepackage{ijcai21}

\usepackage{times}

\usepackage{soul}
\usepackage{url}
\usepackage[hidelinks]{hyperref}
\usepackage[utf8]{inputenc}
\usepackage[small]{caption}
\usepackage{graphicx}
\usepackage{amsmath}
\usepackage{booktabs}
\usepackage{multirow}
\usepackage[flushleft]{threeparttable}
\usepackage{mathrsfs}
\usepackage{bm}

\newcommand{\etal}{\textit{et al}.}

\title{Pairwise Half-graph Discrimination: A Simple Graph-level Self-supervised Strategy for Pre-training Graph Neural Networks}

\author{
Pengyong Li$^{1,2}$\footnote{These two authors contributed equally.},  
Jun Wang$^{2\star}$,  
Ziliang Li$^{2,4}$,  
Yixuan Qiao$^{2}$,  
Xianggen Liu$^{1}$,  
Fei Ma$^{3}$,  
Peng Gao$^{2}$,  \\
Sen Song$^{1\dagger}$ \And
Guotong Xie$^{2,5,6}$ \footnote{Corresponding Author.} 
\affiliations
$^1$Department of Biomedical Engineering, Tsinghua University, Beijing, China\\
$^2$Ping An Healthcare Technology, Beijing, China\\
$^3$Chinese Academy of Medical Sciences, Beijing, China\\
$^4$Central University of Finance and Economics, Beijing, China\\
$^5$Ping An Health Cloud Company Limited, Shenzhen, China\\
$^6$Ping An International Smart City Technology Co., Ltd., Shenzhen, China\\
\emails
lipy0628@163.com,
junwang.deeplearning@gmail.com,
\{liziliang908, qiaoyixuan528, gaopeng712, xieguotong\}@pingan.com.cn,
drmafei@126.com,
\{liuxg16, songsen\}@mail.tsinghua.edu.cn
}

\begin{document}
\maketitle

\begin{abstract}
Self-supervised learning has gradually emerged as a powerful technique for graph representation learning. However, transferable, generalizable, and robust representation learning on graph data still remains a challenge for pre-training graph neural networks.
In this paper, we propose a simple and effective self-supervised pre-training strategy, named Pairwise Half-graph Discrimination (PHD), that explicitly pre-trains a graph neural network at graph-level. PHD is designed as a simple binary classification task to discriminate whether two half-graphs come from the same source. Experiments demonstrate that the PHD is an effective pre-training strategy that offers comparable or superior performance on 13 graph classification tasks compared with state-of-the-art strategies, and achieves notable improvements when combined with node-level strategies. Moreover, the visualization of learned representation revealed that PHD strategy indeed empowers the model to learn graph-level knowledge like the molecular scaffold. These results have established PHD as a powerful and effective self-supervised learning strategy in graph-level representation learning. 

\end{abstract}

\section{Introduction}
Graph modeling has recently received broad interest {because of} the increasing number of non-Euclidean data that needs to be analyzed across various areas, including social networks, physics, and bioinformatics~\cite{hamilton2017representation}. The graph neural network (GNN)~\cite{kipf2016semi,velivckovic2017graph,hamilton2017inductive}, a deep learning-based method, has been reported to be a powerful tool for graph representation learning.  However, supervised training of GNN usually requires labor-intensive labeling and relies on domain expert knowledge. 
One way to alleviate the need for large labeled data is to pre-train a GNN on unlabeled data via self-supervised learning, and then transfer the learned model to downstream tasks. This transfer learning {methodology} has achieved great success in nature language process (NLP) and computer vision (CV)~\cite{liu2020self}. But, there is less exploration~\cite{hu2019strategies,hu2020gpt,rong2020self,you2020graph} of pre-training schemes for GNNs {compared to NLP and CV domains}. Current pre-training schemes for GNN mainly focus on unsupervised representation learning~\cite{sun2019infograph}, which learn graph embeddings on a dataset and the embeddings are fed into a  classifier for the downstream task on this same dataset. The state-of-the-art method in unsupervised node and graph classification is contrastive learning. Contrastive learning techniques are used to train an encoder that builds discriminative representations by comparing positive and negative samples to maximize the mutual information~\cite{liu2020self}. Most of them employ the batch-wise positive/negative samples generation for contrastive discrimination, which bring huge computation costs and {unsuitability for pre-training} on large-scale datasets, while large-scale dataset is essential for pre-training.

\begin{figure*}[t]
\centering
\includegraphics[width=0.7\linewidth]
{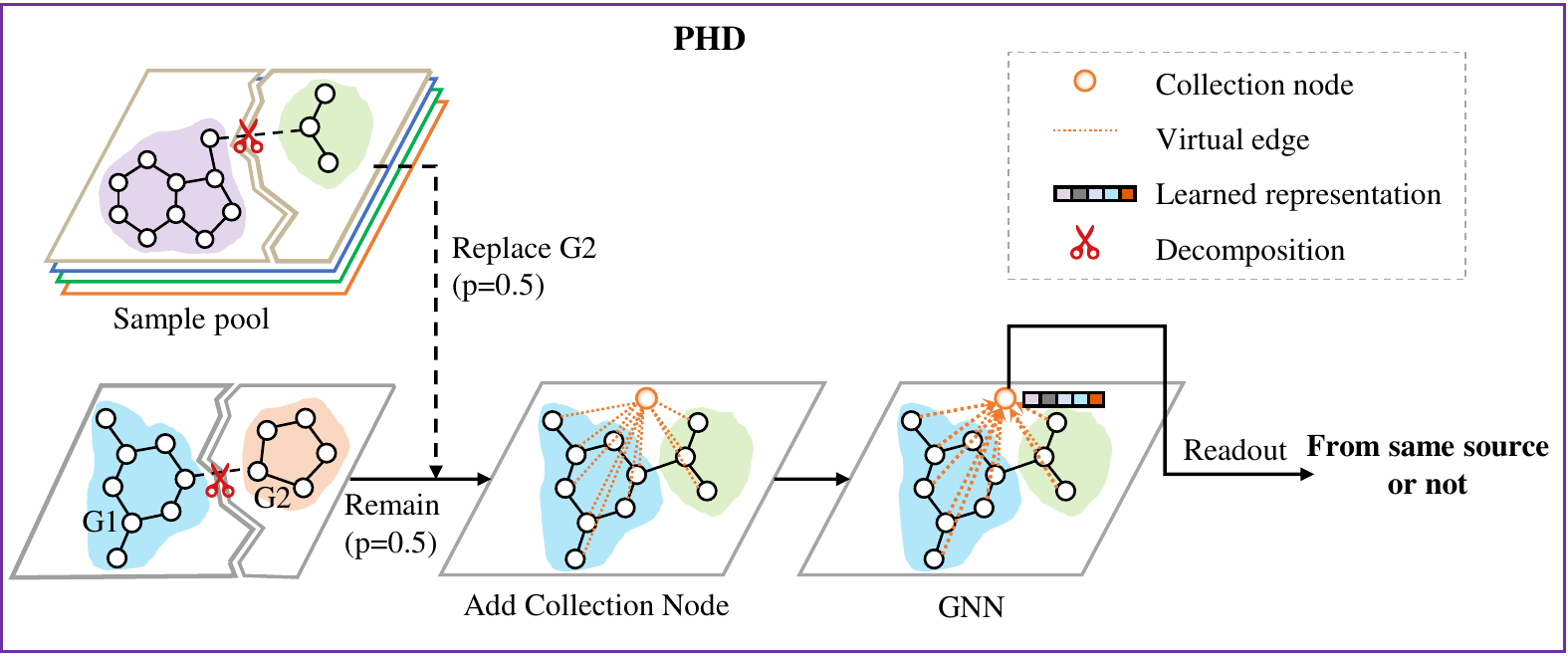}
\caption{Overview of the PHD strategy for pre-training GNNs. PHD task is designed to identify whether two half-graphs come from the same source graph. The graph was firstly decomposed into two half-graphs, one of these two half-graphs has a 0.5 possibility to be replaced by a half-graph from another graph as the negative sample, while the unchanged half-graphs form the positive sample with a 0.5 possibility. Besides, a virtual node called the collection node is added to connect with all other nodes by virtual edges, so as to gather the information from the graph pairs and explicitly learn the graph-level features. Then, each node's hidden states are updated in each GNN message-passing iteration, based on the messages from neighbouring nodes and edges, The binary output is whether two half-graphs are homologous couples.}
\label{fig:architecture}
\end{figure*}


The reason for less works on graph transfer learning might be that some graph datasets are limited in size and GNNs often have shallow architectures~\cite{you2020graph}.  In fact, the graph datasets are recently getting larger (e.g. molecular graph data), and even for shallow GNNs, pre-training could provide a better initialized parameters than random initialization. Moreover, recent researches have proposed many theories and architectures about deep GNNs~\cite{li2020deepergcn,liu2020towards}.  Nowadays, some works have proven the significance of GNN transfer learning. For example, Hu et.al.~\cite{hu2019strategies} and GROVER~\cite{rong2020self} have pre-trained the GNN model on large-scale molecular graph data and achieved impressive performance on multiple molecular properties prediction tasks by transferring the pre-trained model to downstream datasets. However, most of these strategies fell into node-level representation learning, which might not capture global information well and result in limited performance in downstream graph-level tasks. In general, self-supervised learning and pre-training are still insufficiently explored for GNNs.

In this paper, we propose a novel self-supervised strategy, named Pairwise Half-graph Discrimination (PHD), for pre-training GNNs at graph-level. The key idea of PHD is to learn to compare two half-graphs (each decomposed from a graph) and discriminate whether they come from the same source (binary classification). In particular, we employ a virtual node to integrate the information of two half-graphs based on the message passing of GNN. The representation of the virtual node, serving as the global representation of the given two half-graphs, learns to predict the true label of the classification task via maximum likelihood estimation. We conduct a systematic empirical study on the pre-training of various GNN models, to evaluate PHD on transfer learning and unsupervised representation learning. The results demonstrate that PHD is an effective pre-training strategy for graph data.

To summarize, our work presents the following key contributions: 
\begin{itemize}
\item We propose a novel and simple self-supervised strategy named PHD which explicitly pre-trains a GNN at graph-level. 

\item Extensive experiments indicate that PHD achieves superior or comparable results to the state-of-the-art methods on graph transfer learning and unsupervised representation learning.

\item Our PHD can cooperates well with node-level strategies, and can generalize well to different GNN models.

\item We provide a procedure to evaluate whether a self-supervised strategy can empower GNN to learn graph-level knowledge. {By this way, we find PHD can capture intrinsic patterns underlying the graph structures.}

\end{itemize}

\section{Related Works}

\paragraph{Transfer learning.} Transfer learning refers to pre-training a model and re-purposing it on different tasks~\cite{hu2019strategies}. Although transfer learning is a common and effective approach for NLP and CV, it is rarely explored for GNNs, due to the diverse fields that graph-structured data sources are from~\cite{you2020graph}. Current transfer learning schemes on graph data are mainly inspired by language model, such as AttrMasking~\cite{hu2019strategies}, ContextPred~\cite{hu2019strategies}, GPT-GNN~\cite{hu2020gpt}, and GROVER~\cite{rong2020self}. However, most of these strategies fell into node-level representation learning, which might can not capture global information well. Besides the language model inspired methods, some works began to use contrasive learning for transfer learning on graph data, including GraphCL~\cite{you2020graph} and MICRO-Graph~\cite{zhang2020motif}.

\paragraph{Unsupervised learning.} Traditional graph unsupervised learning methods are mainly based on graph kernels. Compared to graph kernel, contrastive learning methods can learn explicit embedding, and achieve better performance, which are the current state-of-the-art in unsupervised node and graph classification tasks. Generally, current contrastive graph learning employs a node-node contrast~\cite{velivckovic2018deep} or node-graph contrast~\cite{sun2019infograph,hassani2020contrastive} to maximize the mutual information.

\section{Methodology}
\subsection{PHD Overview}
\label{sec:PHD}

In order to train a model that captures the global graph information, we pre-train the model with a binarized Pairwise Half-graph Discrimination (PHD) task that can be easily established from any graph database. Simply, PHD task is designed to discriminate whether two half-graphs come from the same source. As shown in Figure \ref{fig:architecture}, the graph is firstly decomposed into two half-graphs, one of these two half-graphs has a 0.5 possibility to be replaced by a half-graph disconnected from another graph which constitutes the negative sample, otherwise the positive sample. Actually, we can regard PHD as a jigsaw puzzle. We assume that two half-graphs decomposed from the same source can be combined into a valid graph, while two half-graphs from the different source can not. By making the model to combine two sub-graph and distinguish whether they can form a valid graph by piecing together, the model can learn the global topological structure in this training process of comparing and combining.

\begin{figure}[ht]
\centering
\includegraphics[width=0.76\linewidth]
{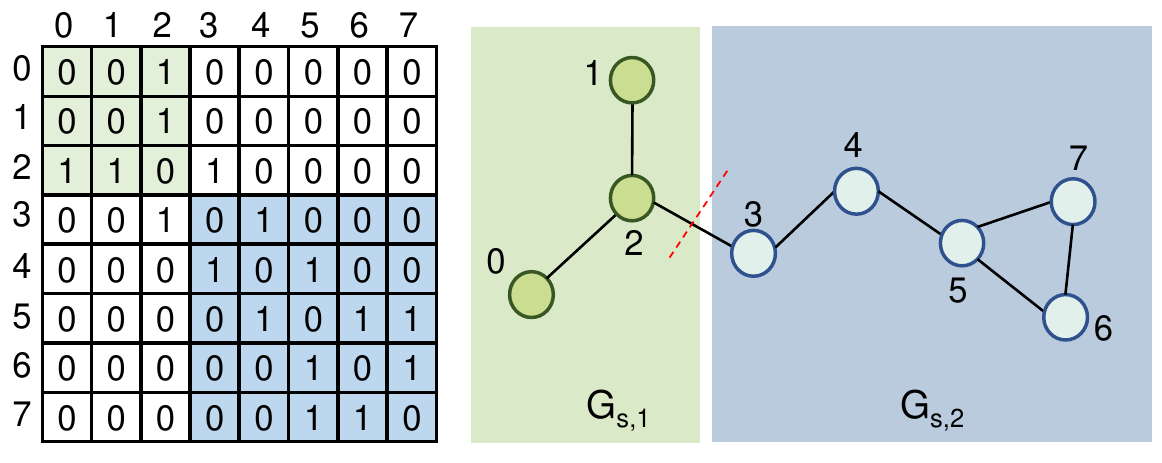}
\caption{The graph decomposition sample. The left sub-figure is the adjacency matrix of the graph in the right sub-figure, where the green and blue represent the decomposed two half-graphs.}
\label{fig:decompo}
\end{figure}

To collect the information from the half-graph pair and learn the graph-level features, a virtual node, called collection node, is added to connect with all other nodes by virtual edges. The embedding $h_c$ of collection node is learned by GNN. Formally, during each message-passing iteration $k$ in a GNN, the vertex update function $U_k$ updates each node's hidden states $h_i$, based on the messages from representations of neighboring nodes and edges.

\begin{align}
    \label{eq:mpnn}
    m_i^{k} &= \sum_{j\in \mathcal{N}_i}\bm M_k(h_i^{k-1},h_j^{k-1},e_{ij}),\\
    h_i^{k} &=  U_k(h_i^{k-1},m_i^{k}),
\end{align}
where $\mathcal{N}_i$ represents the neighbors of node $i$, $M_k$ is the messsage passing function, $e_{ij}$ denotes the edge between the node $i$ and node $j$. And $m_{i}^{k}$ denotes the message node $i$ receives during iteration $k$. 

The message passing runs for $K$ iterations and sequentially updates each node's representation. The final embedding $h_c^K$ of collection node is fed into a simple linear discriminator to make a binary prediction,
\begin{equation}
    \label{eq:overview}
    p=\sigma(\text{NN}(h_c^K))
\end{equation}

where $\sigma$ represents the sigmoid activation function, $\text{NN}$ is a single-layer perceptron network. We employed the cross-entropy loss function for simple computation to optimize the parameters of the network as follows:
\begin{equation}
    L = -\sum_{i=1}^m y\log(p)+(1-y)\log(1-p)
\end{equation}
where $m$ is the number of samples. After pre-trained, the collection node embedding can be regarded as a graph-level representation for the graph and used for downstream tasks. In addition, graph representation can also be obtained by averaging the nodes' embeddings or other global graph pooling methods.

In the following sections, we describe the important components of PHD in detail.

\subsection{Graph Decomposition and Negative Sampling}
We decompose the graph into two half-graphs to generate the half-graph pairs, served as the positive sample, and replace one of the half-graphs to produce the negative sample. As the example shown in Figure \ref{fig:decompo}, given a graph $G=(V,E)$ where $V$ represents nodes and $E$ represents edges. A sampled node $v_3$ is employed as the border node to separate $G$ into two half-graphs $G_{s,1}$ and $G_{s,2}$, where $G_{s,1}$ contains nodes $\{v_0,v_1,v_2\}$ and $G_{s,2}$ contains nodes $\{v_3,v_4,\cdots,v_7\}$. The edges in these two half-graphs correspond to the top-left sub-matrix and bottom-right sub-matrix of the adjacency matrix respectively. In order to produce half-graphs with balanced and various size, the border node index is randomly sampled in the range of 1/3 to 2/3 of the total number of nodes.

For negative sampling, we randomly sample another graph in the dataset and separate it into two half-graphs using the above method, and $G_{s,2}$ is replaced with one of these two half-graphs to generate a negative sample.
How negative samples are generated can have a large impact on the quality of the learned embeddings. It may drive the model to identify whether the two graphs are homologous or estimate whether the two graphs can be combined into a valid graph. In this way, the model can learn the valuable graph-level features of graphs from the nodes and edges which is essential for the downstream tasks.

\subsection{Virtual Collection Node}
The half-graph pair obtained via the above approach are two independent graphs without any connection. 
We concatenate these two half-graphs into a single whole graph, and introduce a virtual collection node to derive the global graph-level representation by aggregating every node information. The collection node is linked with all the other nodes by virtual directed edges, pointing from the other nodes to the collection node. During the message passing process of GNN, the collection node learns its representation from all the other nodes but does not affect the feature update procedure of them. Consequently, the collection node's feature can grasp the global representation of the half-graphs pair and be fed into a feed-forward neural network for the final prediction.

\begin{figure}[ht]
\centering
\includegraphics[width=0.76\linewidth]
{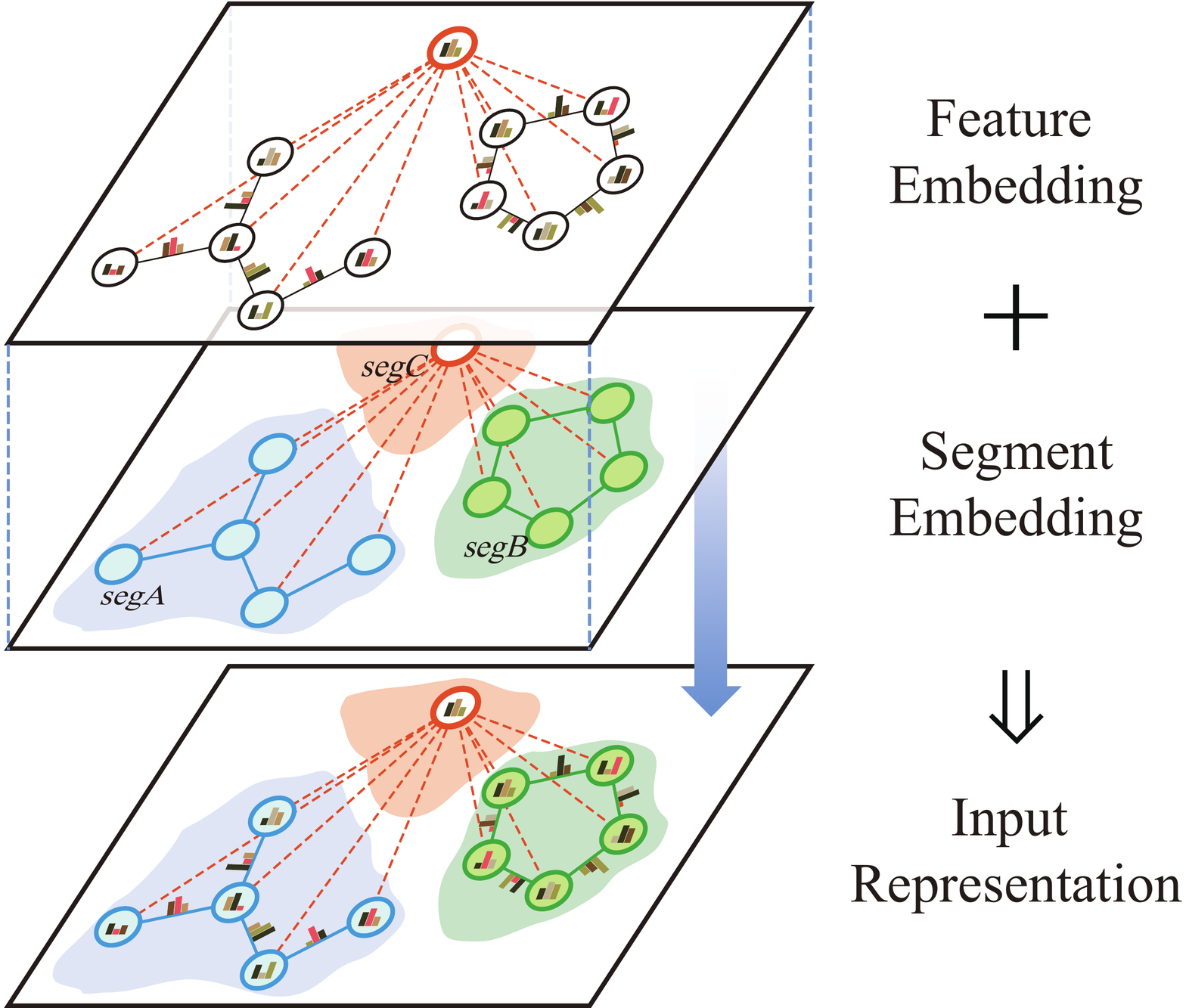}
\caption{The input representation of graph data is constructed by summing two parts: feature embedding and segment embedding. (a) Feature embedding: a set of node and edge features go through the embedding transformation to describe a graph. (b) Segment embedding: a learned segmentation embedding to every node and every edge indicating which half-graph it belongs to,  different colors represents different sementation.}
\label{fig:representation}
\end{figure}

\subsection{Input Representation}
As shown in Figure \ref{fig:representation}, the input representation consists of two parts: feature embedding and segment embedding. A graph is generally described by a set of nodes features and edges features.  Besides the feature embedding, we add a learned segmentation embedding to every node and every edge indicating which half-graph it belongs to. {Specifically, we label each node and edge of  $G_{s,1}$ as 0, each node and edge of  $G_{s,2}$ as 1, collection node and edges connected to collection node as 2. Then these
segment labels are fed into the embedding layer that yields segment embedding. The final input representation is constructed by summing the segment embedding and feature embedding.} In this way, the model could distinguish the nodes and edges from different segments, thus enables simultaneous input of two graphs.

 \begin{table*}[t]
\centering
\resizebox{2\columnwidth}{!}{%
\begin{tabular}{ccccccccccc} \toprule
    \multicolumn{2}{c}{\textbf{Dataset}} & \textbf{BBBP} & \textbf{Tox21} & \textbf{ToxCast} & \textbf{SIDER} & \textbf{ClinTox} & \textbf{MUV} & \textbf{HIV} & \textbf{BACE} & \textbf{Average} \\\midrule
  \multicolumn{2}{c}{No. Molecules}  & 2039  & 7831  &  8575  & 1427  & 1478  & 93087 & 41127 & 1513 & / \\ 
  \multicolumn{2}{c}{No. Binary prediction tasks}  & 1      &  12      & 617      & 27   &  2    & 17  & 1  & 1 & / \\ \midrule
\multicolumn{2}{c}{Pre-training strategy} & \multicolumn{9}{c}{\multirow{2}[2]{*}{Average ROC-AUC across 10 random seeds with scaffold split}} \\ \cmidrule{1-2} 
Graph-level & Node-level & {} \\ \midrule
\multicolumn{2}{c}{No Pre-train} & 63.3 ±2.5 & 75.2 ±0.7 & 64.0 ±0.4 & 53.3 ±2.5 & 62.9 ±2.7 & 71.8 ±1.6 & 75.3 ±1.9 & 74.6 ±2.4 & 67.5 \\ 

-- & Infomax  &  68.8 ±0.8 & 75.3 ±0.5 &  62.7 ±0.4 & 58.4 ±0.8 & 69.9 ±3.0 & 75.3 ±2.5 & 76.0 ±0.7 & 75.9 ±1.6 & 70.3 \\
-- & EdgePred & 67.3 ±2.4 & 76.0 ±0.6 & 64.1 ±0.6 & 60.4 ±0.7 & 64.1 ±3.7 & 74.1 ±2.1 & 76.3 ±1.0 & \bf{79.9 ±0.9} & 70.3 \\ 
-- & ContextPred  &  68.0 ±2.0 &  75.7 ±0.7 & 63.9 ±0.6 & 60.9 ±0.6 & 65.9 ±3.8 & 75.8 ±1.7 & 77.3 ±1.0 & 79.6 ±1.2 & 70.9 \\
-- & AttrMasking  &  64.3 ±2.8 & 76.7 ±0.4 & 64.2 ±0.5 & 61.0 ±0.7 & 71.8 ±4.1 & 74.7 ±1.4 & 77.2 ±1.1 & 79.3 ±1.6  & 71.1 \\ 
GraphCL &--& \bf{69.7 ±0.7} & 73.9 ±0.7 & 62.4 ±0.6 & 60.5 ±0.9 & \bf{76.0 ±2.7} & 69.8 ±2.7 & 78.5 ±1.2 & 75.4 ±1.4 & 70.7 \\ 

bi-PHD & --  & 66.8 ±2.2 & \textbf{78.1 ±0.6} & \textbf{66.4 ±0.3} & \textbf{64.5 ±1.2} & 65.7±1.8 & 75.4 ±1.4 & 77.0±0.5 & {79.3 ±0.8} & 71.6 \\

PHD   & --     &68.8 ±0.7 &77.9 ±0.7 &\bf{ 66.4 ±0.6} & 60.4 ±0.6 &69.5 ±3.8  & \bf{76.5 ±2.0 }& \bf{78.6 ±1.3} & 77.5 ±2.0 & \bf{71.9} \\ \hline 
PHD & ContextPred  &   {\bf 69.7 ±2.1} & { 77.0 ±0.6} &{\bf 66.9 ±0.4 }&{ 61.0 ±0.5} & {69.8 ±3.0} & { 76.1 ±1.2} & {78.3 ±0.9} & {\bf 81.3 ±1.9} & {72.5} \\ 
PHD & AttrMasking  &{69.2 ±0.9} & { 77.8 ±0.5} &{66.8 ±0.7 }& {\bf 62.8 ±1.3} & {\bf 74.8 ±4.7} & { 75.2 ±1.2} & {\bf79.1 ±0.9} & {79.7 ±1.8 }& {\bf73.2} \\ 
\bottomrule
\end{tabular}
}
\vspace{0.0cm}
 \caption{ Test ROC-AUC score of different pre-training strategies on transfer learning (following the same experimental setting as [Hu \textit{et al.}, 2019], pre-training the GIN model on 2 million molecules sampled from the ZINC15). bi-PHD represents the PHD strategy with the bidirectional edges between the collection node and the rest nodes.}
 \label{table:result_molecule}
 \vspace{-0.2cm}
\end{table*}

\section{Experiments on Transfer Learning}
\label{sec:exp}

We performed experiments on transfer learning on molecular property prediction following \cite{hu2019strategies,you2020graph}, which pre-trains GNNs on a large-scale molecular graph data and finetunes the model in different datasets to evaluate the transferability of the pre-training scheme.
\subsection{Datasets}
For pre-training, our PHD is performed on 2 million unlabeled molecules sampled from the ZINC15~\cite{sterling2015zinc} database. After pre-training, we fine-tuned the model on 8 downstream task datasets including MUV, HIV, BACE, BBBP, Tox21, ToxCast, SIDER, and ClinTox. All the molecules in these datasets are described by a set of node and bond features obtained by RDKit as same as Hu \etal~\cite{hu2019strategies}.

\subsection{Experimental Setup}
We evaluate the effectiveness of PHD from three perspective on large-scale graph data: 1) Whether PHD can pre-train better GNNs that generalize well on graph classification tasks; 2)Whether our graph-level strategy---PHD can cooperate well with node-level strategies; 3)Whether PHD can generalize well to different GNNs.

For perspective 1), we systematically compared PHD with some strong baselines including  Infomax~\cite{velivckovic2018deep}, EdgePred~\cite{hamilton2017representation}, AttrMasking~\cite{hu2019strategies}, ContextPred~\cite{hu2019strategies}, and GraphCL~\cite{you2020graph}. For perspective 2), we combined PHD with the node-level strategies---AttrMasking and ContextPred to test whether the performances are improved. For perspective 3), we chose four popular GNN architectures including GIN~\cite{xu2018powerful}, GCN~\cite{kipf2016semi}, GAT~\cite{velivckovic2017graph} and GraphSAGE~\cite{hamilton2017inductive} to evaluate PHD.

\subsection{Experiment Configuration}
We adopt same experimental setting as Hu \etal~\cite{hu2019strategies}. The different GNN architectures including GIN, GCN, GAT, and GraphSAGE were adapted from the implementation in Hu \etal~\cite{hu2019strategies}. All the GNN architectures have 5 GNN message passing layers with 300 embedding dimension. We run all pre-training methods for 100 epochs with a learning rate of 0.001 and a batch size of 256 on 2 million molecules from ZINC15. After pre-training, we add a linear classifier to fine-tune on 8 datasets above. We split these datasets via \textit{scaffold split}~\cite{hu2019strategies}  with the ratio of 8:1:1 (train:validation:test). We train models for 100 epochs with a learning rate of 0.001 and the dropout rate of 50\%. The validation sets were used for model selection and the ROC-AUC scores on test sets are reported. We report the mean ROC-AUC and standard deviation of experiments with 10 random seeds. We report results from previous papers with the same experimental setup if available. 
 
\begin{table*}[ht]
\center
\scalebox{0.9}{
\begin{tabular}{@{}ccccccccc@{}}
\toprule
\multicolumn{1}{c}{\begin{tabular}[c]{@{}c@{}} Dataset \end{tabular}} & \multicolumn{0}{c}{\textbf{BBBP}} & \textbf{Tox21} & \textbf{ToxCast} & \textbf{SIDER} & \textbf{ClinTox} & \textbf{HIV} & \textbf{BACE} & \textbf{Average}\\ \midrule
No Pre-train & 82.1 $\pm$ 1.7 & 76.1 $\pm$ 0.6 & 63.3 $\pm$ 0.8 & 55.7 $\pm$ 1.4 & 75.0 $\pm$ 3.6 & 73.4 $\pm$ 0.9 & 72.8 $\pm$ 2.1 & 71.2 \\ \midrule
InfoGraph & 80.4 $\pm$ 1.2 & 76.1 $\pm$ 1.1 & 64.4 $\pm$ 0.8 & 56.9 $\pm$ 1.8 & \textbf{78.4 $\pm$ 4.0} & 72.6 $\pm$ 1.0 & 76.1 $\pm$ 1.6 & 72.1 (+0.9\%)\\
GPT-GNN & 83.4 $\pm$ 1.7 & 76.3 $\pm$ 0.7 & 64.8 $\pm$ 0.6 & 55.6 $\pm$ 1.6 & 74.8 $\pm$ 3.5 & 74.8 $\pm$ 1.0 & 75.6 $\pm$ 2.5 & 72.2 (+1.0\%)\\
GROVER & 83.2 $\pm$ 1.4 & 76.8 $\pm$ 0.8 & 64.4 $\pm$ 0.8 & 56.6 $\pm$ 1.5 & 76.8 $\pm$ 3.3 & 74.5 $\pm$ 1.0 & 75.2 $\pm$ 2.3 & 72.5 (+1.3\%)\\
MICRO-Graph & 83.8 $\pm$ 1.8 & 76.7 $\pm$ 0.4& 65.4 $\pm$ 0.6 & 57.3 $\pm$ 1.1 & 77.5 $\pm$ 3.4 & 75.5 $\pm$ 0.7 & 76.2 $\pm$ 2.5 & 73.2 (+2.0\%)\\
\midrule
PHD & \textbf{86.0 $\pm$ 0.7} & \textbf{82.9 $\pm$ 0.2} & \textbf{72.7 $\pm$ 0.2} & \textbf{61.0 $\pm$ 0.5} & 74.3 $\pm$ 1.2 & \textbf{79.6 $\pm$ 0.6} & \textbf{80.6 $\pm$ 0.9} & \textbf{76.7 (+5.5\%)}\\
\bottomrule
\end{tabular}}
\caption{10-fold cross validation ROC-AUC score comparison results (following the same experimental setting as MICRO-Graph, which pre-trained the DeeperGCN model on the HIV dataset, and fine-tuned the pre-trained model on each downstream tasks. The results of baselines are taken from  MICRO-Graph. )}
\label{tab:transfer}
\end{table*}

  \begin{table*}[t]
\centering
\resizebox{2\columnwidth}{!}{%
    \begin{tabular}{clcccccccccc}
    \toprule
    \textbf{Architecture} & \textbf{Pretrain strategy} & \textbf{BBBP} & \textbf{Tox21} & \textbf{ToxCast} & \textbf{SIDER} & \textbf{ClinTox} & \textbf{MUV} & \textbf{HIV} & \textbf{BACE} & \textbf{Average} \\
 \midrule
    \multirow{3}[1]{*}{\textbf{GCN}} &No Pre-train  & 66.5 ±0.5 & 74.9 ±0.4 & 64.2 ±0.4 & \textbf{58.2 ±0.4} & 55.2 ±0.9 & 72.3 ±0.9 & \textbf{73.2 ±1.6} & 76.3 ±0.7 & 67.6 \\
          & PHD    & \textbf{66.7 ±0.6} & \textbf{75.9 ±0.3} & \textbf{64.8 ±0.3} & 56.7 ±0.5 & \textbf{64.2 ±1.8} & 69.5 ±0.7 & 69.7 ±1.6 & 77.4 ±0.6 & 68.1 \\
          & PHD + node-level & 66.1 ±1.6 & 74.5 ±0.4 & 64.6 ±0.2 & 57.0 ±0.2 & 62.5 ±0.6 & \textbf{72.6 ±1.2} & 71.4 ±0.9 & \textbf{77.6 ±0.9} & \textbf{68.2} \\
    \midrule
    \multirow{3}[2]{*}{\textbf{GAT}} & No Pre-train     & 65.4 ±1.2 & 71.8 ±0.6 & 57.4 ±0.6 & \textbf{60.3 ±1.0} & 62.0 ±3.1 & \textbf{67.1 ±1.1} & \textbf{73.9 ±1.9} & 67.4 ±2.8 & 65.6 \\
          & PHD        & 64.7 ±0.7 & 71.9 ±0.5 & 62.7 ±0.6 & 55.1 ±1.3 & 65.0 ±2.1 & 66.4 ±0.8 & 73.3 ±1.1 & 70.6 ±1.2 & 66.2 \\
          & PHD + node-level & \textbf{66.2 ±1.3} & \textbf{74.5 ±0.6} & \textbf{64.1 ±0.4} & 57.5 ±2.4 & \textbf{69.4 ±1.9} & 66.8 ±0.9 & 72.4 ±0.8 & \textbf{73.1 ±1.3} & \textbf{67.9} \\
    \midrule
    \multirow{3}[1]{*}{\textbf{GraphSAGE}} &No Pre-train    & 68.6 ±1.5 & 75.4 ±0.5 & 64.7 ±0.6 & 58.8 ±1.4 & 59.5 ±2.9 & 74.2 ±1.0 & 75.9 ±1.1 & 65.1 ±1.0 & 67.8 \\
          & PHD    & 66.3 ±1.1 & 75.9 ±0.4 & 64.8 ±0.4 & 61.1 ±0.6 & \textbf{64.3 ±1.8} & \textbf{77.8 ±1.3} & 77.3 ±0.5 & 76.6 ±0.6 & 70.5 \\
          & PHD + node-level & \textbf{71.7 ±0.9} & \textbf{76.1 ±0.4} & \textbf{65.5 ±0.4} & \textbf{62.7 ±0.8} & 60.0 ±3.7 & 76.5 ±0.8 & \textbf{78.1 ±0.7} & \textbf{80.1 ±0.7} & \textbf{71.3} \\ \bottomrule
    \end{tabular}%
    }
 
 \caption{Test ROC-AUC score performances on different GNN architectures. ContextPred is used as node-level strategy. The best results for each dataset and comparable results are in bold. }
 \label{table:result_compare}
 \vspace{-0.2cm}
\end{table*}%

 \subsection{Experimental Results}
 
 \subsubsection{Whether PHD Can Pre-train Better GNN?}
 The upper half of Table \ref{table:result_molecule} shows that our PHD strategy gives better predictive performance than the previous strategies in terms of the average ROC-AUC score on eight datasets. For example, on Toxcast, we observe 3.4\% relative improvement over previous state-of-the-art. These results indicate that leveraging global structural information plays a more essential role in self-supervised graph learning. In particular, the edges between the collection node and other nodes in our PHD strategy can be unidirectional or bidirectional. We conduct the ablation study of the edges direction. Table \ref{table:result_molecule} shows that unidirectional edges outperform bidirectional edges (bi-PHD) in terms of average ROC-AUC. What is more, the bidirectional edges need more computation than unidirectional edges, so we employed the unidirectional message passing for collection node in all our experiments.

 Aside from above methods, we also compare with the other four recent proposed strategies as InfoGraph~\cite{sun2019infograph}, GPT-GNN~\cite{hu2020gpt}, GROVER~\cite{rong2020self}, and MICRO-Graph~\cite{zhang2020motif}. Because InfoGraph has a huge computation cost that unsuited to pre-train on large-scale dataset, we pre-trained on a small-scale dataset following the same experimental setting as MICRO-Graph. Specifically, we pre-trained DeeperGCN~\cite{li2020deepergcn} on the HIV dataset which contains 40K molecules. Then we fine-tuned the pre-trained model on downstream tasks. The best 10-fold cross validation ROC-AUC scores averaged over 10 folds are reported in Table \ref{tab:transfer}.  Table \ref{tab:transfer} shows that PHD outperform previous best schemes on 6 of 7 datasets by a large margin (about 5.5\% average improvement).

\subsubsection{Whether PHD Can Cooperate Well with Node-level Strategies?}
In general, to pre-train well on graph data, we need to pre-train a model at both graph and node-level to encode more information\cite{hu2019strategies}. 
Lower half of table \ref{table:result_molecule} summarizes the results of combing our PHD with node-level strategies. It shows that the combinations significantly outperform the strategies of only PHD or node-level strategy. Furthermore, the PHD combining with AttrMasking gives a better predictive performance than that with ContextPred, achieving state-of-the-art performance. These results reveal that combining PHD with node-level strategy has the potential of pre-training a GNN model well on graph data, especially on molecular representation learning.

\subsubsection{Whether PHD Can Generalize Well to Different GNNs?}
We conducted experiments with other popular GNN architectures: GCN, GAT, and GraphSAGE. Table \ref{table:result_compare} shows that PHD pre-training yields superior performance than random initialization over different GNN architectures on most datasets, and the performance could be further enhanced when cooperated with node-level strategy. Thus, PHD is architecture-free that could empower different GNN architectures to learn an expressive graph representation. It provides a favorable initialization for model and improves the performance of downstream tasks.  Although our PHD strategy acts universally well to enhance many GNN-based models, different GNN architectures have different gains after pre-trained. Our experiments indicate that GraphSAGE and GIN achieve larger improvement than GCN and GAT. It remains an open challenge in research to explore the relationship between the pre-training strategies and the GNN architectures.

\section{Experiments on Unsupervised Learning}

\begin{table}[t!]
	\centering
	\resizebox{1.0\columnwidth}{!}{%
	\begin{tabular}{cccccccccccc}
	    \toprule
		\multicolumn{2}{c}{\textbf{Dataset}} &      \multicolumn{1}{c}{\textbf{MUTAG}} &	\multicolumn{1}{c}{\textbf{PTC-MR}} &  
		\multicolumn{1}{c}{\textbf{RDT-B}} &  	 \multicolumn{1}{c}{\textbf{IMDB-B}} &	 \multicolumn{1}{c}{\textbf{IMDB-M}} \\
		\midrule
	
		
		\multicolumn{2}{c}{\textbf{(No. Graphs)}} &     {188} &	 {344}  &  {2000}  &	 {1000} &	 {1500} \\  
		
		\multicolumn{2}{c}{\textbf{(No. classes)}} &    {2} &	 {2}  &  {2}  &	 {2} &	 {3} \\  
		
		\multicolumn{2}{c}{\textbf{(Avg. Nodes)}}    &  {17.93} &
		{14.29}  &  {429.63} &  {19.77} &	 {13.00} \\ 
	\midrule
	 & GLK     &81.7 ±2.1  & 57.3 ±1.0  & 77.3 ±0.2 & 65.9 ±1.0  & 43.9 ±0.4  
   \\  
		&WL  &80.7 ±3.0  & 58.0 ±0.9  & 68.8 ±0.4 & 72.3 ±3.4  & 47.0 ±0.6  
   \\  
		&DGK   &87.4 ±0.9  & 60.1 ±1.5  & 78.0 ±1.0 & 67.0 ±0.6  & 44.6 ±0.5 
   \\	
		&sub2vec     &61.1 ±0.6  & 60.0 ±1.1 & 71.5 ±1.0 & 55.3 ±0.8  & 36.7 ±0.7 \\
		&graph2vec  &83.2 ±1.7  & 60.2 ±1.3  & 75.8 ±0.8    & 71.1 ±0.9  & 50.4 ±0.8
   \\

        &InfoGraph  &89.0 ±1.1 & 61.6 ±1.8  & 82.5 ±1.4 & 73.0 ±0.9  & 49.7 ±0.8 
   \\ 
           &GraphCL  &86.8 ±1.3 & --  & \textbf{89.5 ±0.9} & 71.1 ±0.4  & -- 
   \\ 
        &CMV  &89.7 ±1.1  &62.5 ±1.7  & 84.5 ±0.6 & \textbf{74.2 ±0.7} & \textbf{51.2 ±0.5} \\
\midrule
		&\textbf{PHD} & \textbf{90.5 ±0.9} &\textbf{ 65.5 ±0.8}  & \textbf{89.2 ±0.4} & 72.5 ±0.3  & 49.8 ±0.7\\

  
 \bottomrule

	\end{tabular}
	}

\caption{Comparison of 10-fold cross validation classification accuracy on unsupervised learning over 5 datasets. The result in bold indicates the best reported classification accuracy. } 
    \label{table:social}
	
\end{table}

Besides transfer learning, we then evaluated PHD in the unsupervised representation learning, where unsupervised methods gengerate graph embeddings that are fed into a linear SVM classifier.
We conduct experiments on five small-scale benchmark datasets: MUTAG, PTC-MR, REDDIT-BINARY, IMDB-BINARY, and IMDB-MULTI, to compare PHD with previous state-of-the-art approaches including graph kernel methods---GLK~\cite{shervashidze2009efficient}, WL~\cite{shervashidze2011weisfeiler}, DGK~\cite{yanardag2015deep}, and some other unsupervised graph learning methods---sub2vec~\cite{sub2vec}, graph2vec~\cite{narayanan2017graph2vec}, InfoGraph~\cite{sun2019infograph}, GraphCL~\cite{you2020graph}, and Contrastive   Multi-View (CMV)~\cite{hassani2020contrastive}. 

We adopt the same procedure of previous works~\cite{sun2019infograph,you2020graph,hassani2020contrastive} to make a fair comparison, and used 10-fold cross validation accuracy to report the classification performance. Specifically, we first pre-trained GIN model implemented by InfoGraph. The embedding dimension is set to 512. The graph representation learned by the GIN model was fed into a linear SVM to obtain the final prediction. The \textit{C} parameter of SVM was selected from $\{10^{-3},10^{-2},\cdots,10^2,10^3\}$.

Table \ref{table:social} shows that PHD still achieves state-of-the-art performance on MUTAG and PTC-MR and comparable performance on the rest datasets. The great improvement on MUTAG and PTC-MR indicates that PHD is good at handling the small attributed graphs. 

 \section{Explainable Representation Visualization}
 To intuitively prove our PHD is a graph-level strategy, we visualized the representation extracted by the pre-trained models. The primary problem to be solved is to define what can represent global knowledge in a graph.  Fortunately, the scaffold concept in chemistry is proposed to represent the core structures of a molecule, which reveals the graph-level intrinsic patterns to some extent~\cite{bemis1996properties}.  Here, we employed UMAP~\cite{mcinnes2018umap} for molecular graph representation visualization coloring with the corresponding scaffold.  Specifically, we chose 9 most common scaffolds from ZINC dataset, and randomly sampled 1000 molecules from each selected scaffold. Finally, we have 9000 molecules labeled with nine different scaffolds. The molecular representations are obtained by averaging the node embeddings extracted by the last layer of a GIN model, we compared the UMAP visualization of representation results including (a) no pre-trained, (b) PHD, (c) AttrMasking and (d) combining AttrMasking with PHD. 
 
\begin{figure}[ht]
\centering
\includegraphics[width=0.95\linewidth]
{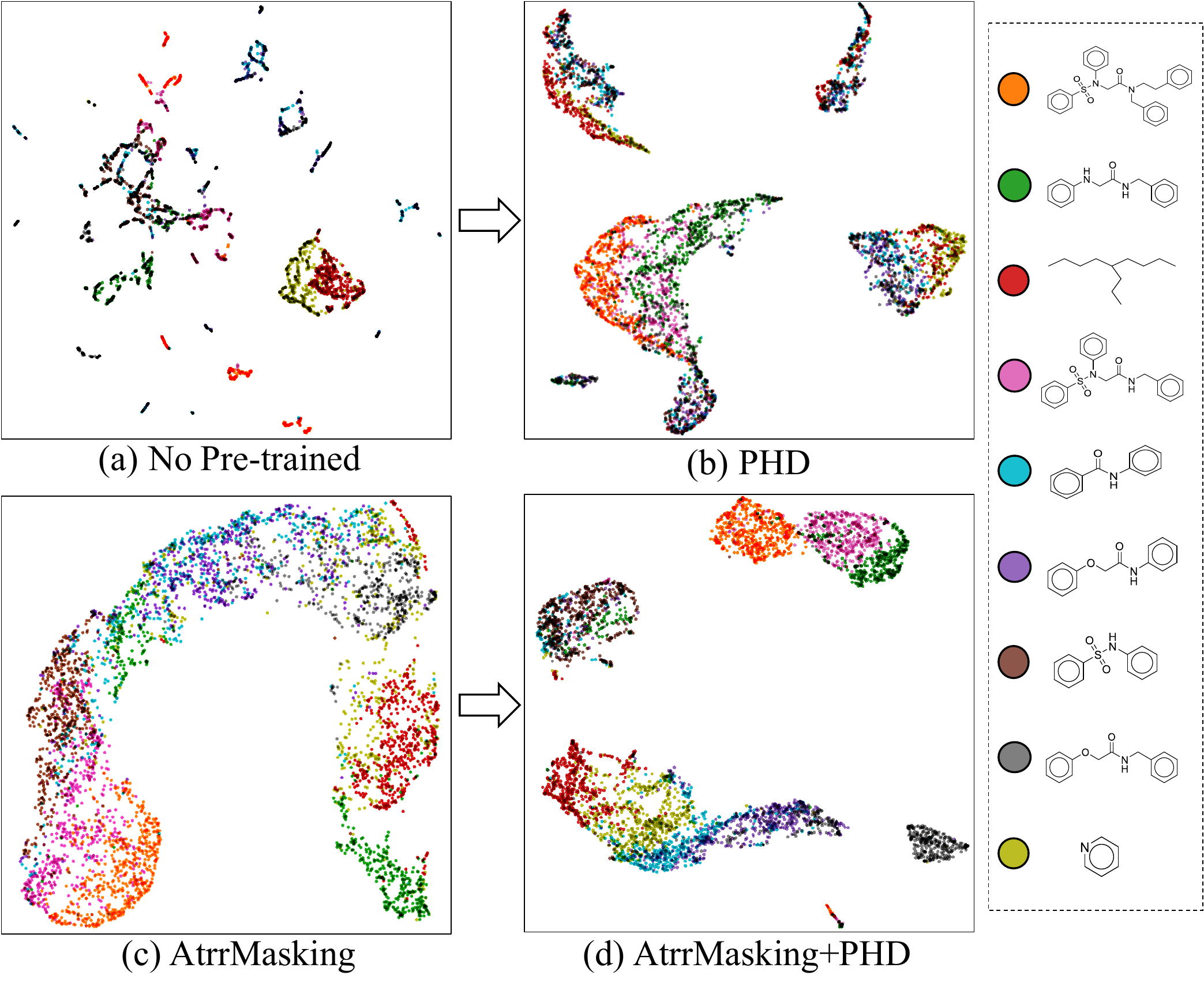}\caption{UMAP visualization of representation learned from the GIN model with or without PHD pre-training: (a) No pre-trained, (b) With PHD pre-training, (c) With Node-level(AttrMasking) pre-training, (d) With Node-level(AttrMasking) + PHD pre-training.}
\label{fig:visual}
\end{figure}

 From Figure \ref{fig:visual}, we observe that the no pre-trained GNN model (a) shows no obvious cluster and the molecules overlap in a mess without meaningful patterns. In contrast, the GNN model pre-trained with our PHD strategy (b) exhibits a discernible clustering. Moreover, although the model pre-trained with node-level AttrMasking (c) scatters the molecules with slight cluster, it still presents notable overlap. Alternatively, with a beneficial combination of PHD and node-level strategy, the representations extracted from the pre-trained model (d) exhibit a more favorable and distinctive clustering corresponding to the nine scaffolds. These results indicate that our PHD strategy prompts the pre-training model to better capture data's globally inherent characteristics, which provides high-quality representations for the downstream graph-level tasks.

\vspace*{-12pt}

 \section{Conclusions}

In this work, we present a self-supervised strategy named Pairwise Half-graph Discrimination (PHD), an effective and simple strategy that explicitly pre-trains the expressive GNN at graph level. Extensive experiments on multiple downstream benchmarks show that the PHD achieves superior performance than state-of-the-art self-supervised strategies on transfer learning and unsupervised representation learning. Moreover, we observe that the pre-trained GNN model with PHD strategy can capture the global graph-level knowledge like the molecular scaffold. In the future, we plan to use PHD for pre-training more expressive GNN models on larger datasets and evaluate it on more downstream tasks.

\bibliographystyle{named}
\bibliography{ijcai21}

\end{document}